\documentclass[conference]{IEEEtran}
\IEEEoverridecommandlockouts
\usepackage{cite}
\usepackage{amsmath,amssymb,amsfonts}
\usepackage{algorithmic}
\usepackage{graphicx}
\usepackage{textcomp}
\usepackage[table,xcdraw]{xcolor} % 支持表格颜色
\usepackage{booktabs} % 用于美化表格线条
\usepackage{xcolor}
\usepackage{booktabs} % 用于美化表格线条和加粗线
\usepackage{colortbl} % 处理表格背景色
\usepackage{multirow}
\usepackage{graphicx} % 在导言区加载宏包
\usepackage[utf8]{inputenc}
\usepackage{authblk}

\def\BibTeX{{\rm B\kern-.05em{\sc i\kern-.025em b}\kern-.08em
    T\kern-.1667em\lower.7ex\hbox{E}\kern-.125emX}}
\begin{document}

\title{FMNet: Frequency-Assisted Mamba-Like Linear Attention Network for Camouflaged Object Detection}

% \author[1]{Ming Deng}
% \author[2,4]{Sijin Sun}
% \author[1]{Zihao Li}
% \author[3]{Xiaochuan Hu}
% \author[1,*]{Xing Wu}

% \affil[1]{Shanghai University}
% \affil[2]{National University of Singapore}
% \affil[3]{University of Electronic Science and Technology of China}

\author{
\textbf{Ming Deng}$^{1,*}$ \hspace{10pt}
\textbf{Sijin Sun}$^{2,4,*}$ \hspace{10pt}
\textbf{Zihao Li}$^{1}$ \hspace{10pt}
\textbf{Xiaochuan Hu}$^{3}$ \hspace{10pt}
\textbf{Xing Wu}$^{1,\dagger}$
}
\affil{\normalsize
$^1$ Shanghai University \quad
$^2$ Agency for Science, Technology and Research \\
$^3$ University of Electronic Science and Technology of China \quad
$^4$ National University of Singapore \\
\footnotesize{$^*$Equal contribution \quad $^\dagger$Corresponding author} 
}

\date{\today}

\maketitle

\begin{abstract}
%伪装目标检测（COD） 由于目标与其周围环境高度相似，导致识别极具挑战性。现有方法主要依赖空间局部特征，难以捕获全局信息，而 Transformer 虽然有效，但会显著增加计算成本。为解决这一问题，我们提出了频率辅助类 Mamba 线性注意力网络（FMNet），利用频域学习高效捕获全局特征，缓解隐藏目标与背景之间的歧义。FMNet 引入了多尺度频率辅助类 Mamba 线性注意力（MFM）模块，将频域与空间特征融合，降低计算复杂度。此外，设计了金字塔频率注意力提取（PFAE）模块和频率反向解码器（FRD），用于语义增强与特征重建。实验结果表明，FMNet 在多个 COD 数据集上表现优异，性能与效率均超过现有方法。代码：https://anonymous.4open.science/r/FMNet-3CE5。
Camouflaged Object Detection (COD) is challenging due to the strong similarity between camouflaged objects and their surroundings, which complicates identification. Existing methods mainly rely on spatial local features, failing to capture global information, while Transformers increase computational costs. To address this, the Frequency-Assisted Mamba-Like Linear Attention Network (FMNet) is proposed, which leverages frequency-domain learning to efficiently capture global features and mitigate ambiguity between objects and the background. FMNet introduces the Multi-Scale Frequency-Assisted Mamba-Like Linear Attention (MFM) module, integrating frequency and spatial features through a multi-scale structure to handle scale variations while reducing computational complexity. Additionally, the Pyramidal Frequency Attention Extraction (PFAE) module and the Frequency Reverse Decoder (FRD) enhance semantics and reconstruct features. Experimental results demonstrate that FMNet outperforms existing methods on multiple COD datasets, showcasing its advantages in both performance and efficiency. Code available at https://github.com/Chranos/FMNet.

% The source code is available at: https://github.com/Chranos/FMNet
\end{abstract}


\section{Introduction}
\label{sec:intro}

% 伪装目标检测（COD）是一项极具挑战性的任务，目的是准确检测那些与周围环境高度相似的隐藏目标，一些生物如变色龙，毛虫等通过“伪装”机制融入自然环境以躲避天敌，COD也是诸多领域的挑战，如医学的肿瘤检测，工业的缺陷检测以及军事方面的运用。
Camouflaged Object Detection (COD) is a highly challenging task aimed at accurately detecting hidden objects that closely resemble their surrounding environment. COD has significant application value in various fields, such as lesion segmentation in medical imaging and defect detection in industrial settings.
% 传统的COD方法依赖于手工特征提取，虽然在特定的场景中表现良好，但是能力有限。随着COD数据集的开源，基于深度学习的COD方法自动提取丰富的特征，有着明显的优势。近年的研究针对特征信息的优化进行探索，为了解决伪装目标大小不一，环境遮挡，边界模糊等问题，多尺度，边界引导，感受野增强的策略等被提出。尽管这些模型取得了一些良好的效果，一些方法往往只是优化了局部特征，难以获取全局信息，这对于大小不一和被遮挡的伪装目标识别不利，另外，空间特征容易受到背景的干扰，这类问题往往是因为空间信息强调局部和单一像素的位置导致的。

Traditional COD methods primarily rely on handcrafted feature extraction\cite{tra1}\cite{tra2}. While these methods achieve certain performance in specific scenarios, their robustness is limited. With the availability of open-source COD datasets\cite{COD10K}\cite{CAMO}\cite{Nc4k}, deep learning-based COD methods have shown significant advantages by automatically extracting rich features\cite{COD10K}\cite{codassplike3}. Recent studies have demonstrated that optimizing feature representation can effectively address challenges such as varying target sizes, occlusion by surroundings, and blurred boundaries in camouflaged objects. Strategies such as multi-scale feature extraction\cite{COD10KV3}\cite{ZoomNet} and boundary guidance\cite{BoundaryGuidedCO} have been proposed. However, most existing methods are still constrained to optimizing local features, making it difficult to capture global information effectively. This limitation is particularly pronounced when detecting camouflaged objects with significant size variations or occlusions. Additionally, spatial domain features are prone to interference from complex backgrounds, often caused by overemphasizing local details or individual pixel positions.

%Transfomer方法被用于获取长距离关系，通过傅立叶变换生成的频域特征也表现出全局特征，这些都有利于网络进行全局上下文建模，然而VIT，PVT之类的视觉Transfomer方法以及频域的转换会导致网络结构参数量增大，带来巨大的计算开销。近年来，Linear Attention， Mamba都用以解决Transfomer复杂度高的问题，基于Mamba的方法在伪装目标检测上少有试验。

\begin{figure}[htbp]
\centering
\includegraphics[width=0.4\textwidth]{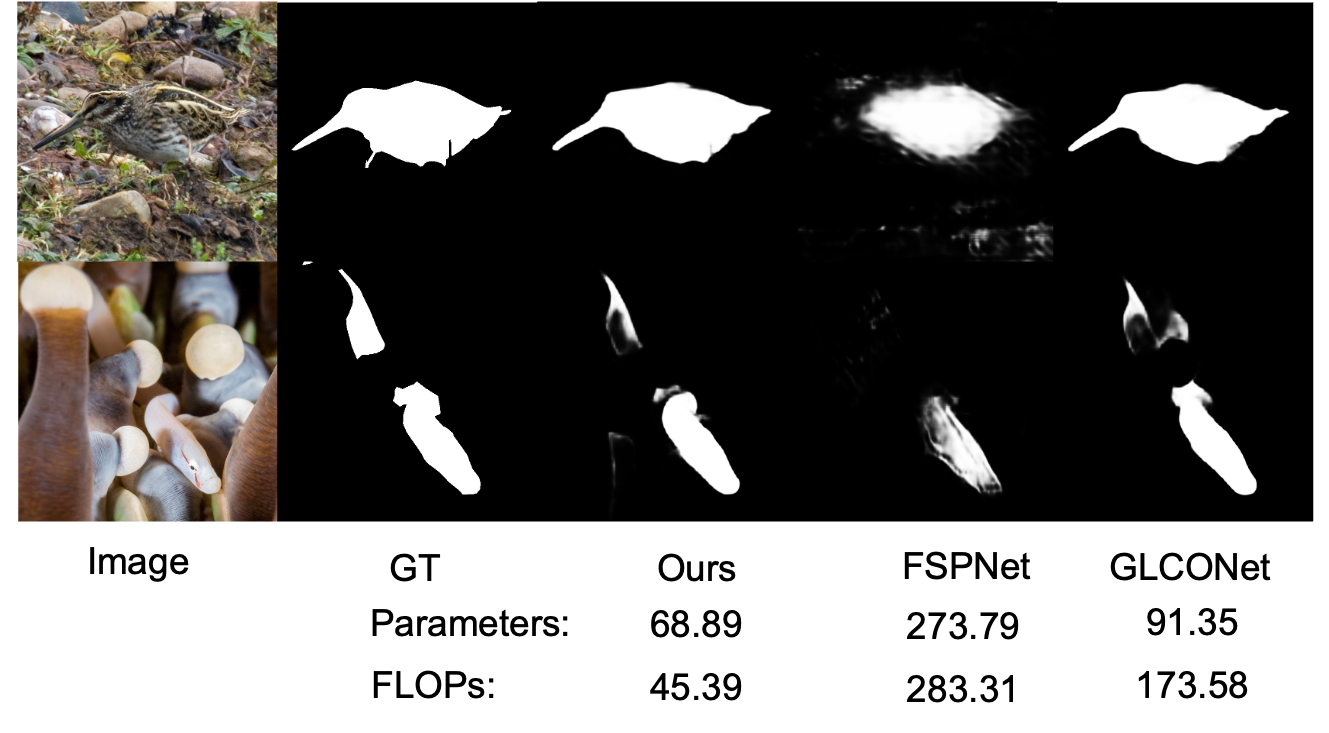}
\caption{Comparison of the proposed method with others using traditional attention mechanisms.}
\label{CT}
\end{figure}

\begin{figure*}[!htbp]
    \centering
    \includegraphics[width=0.8\textwidth]{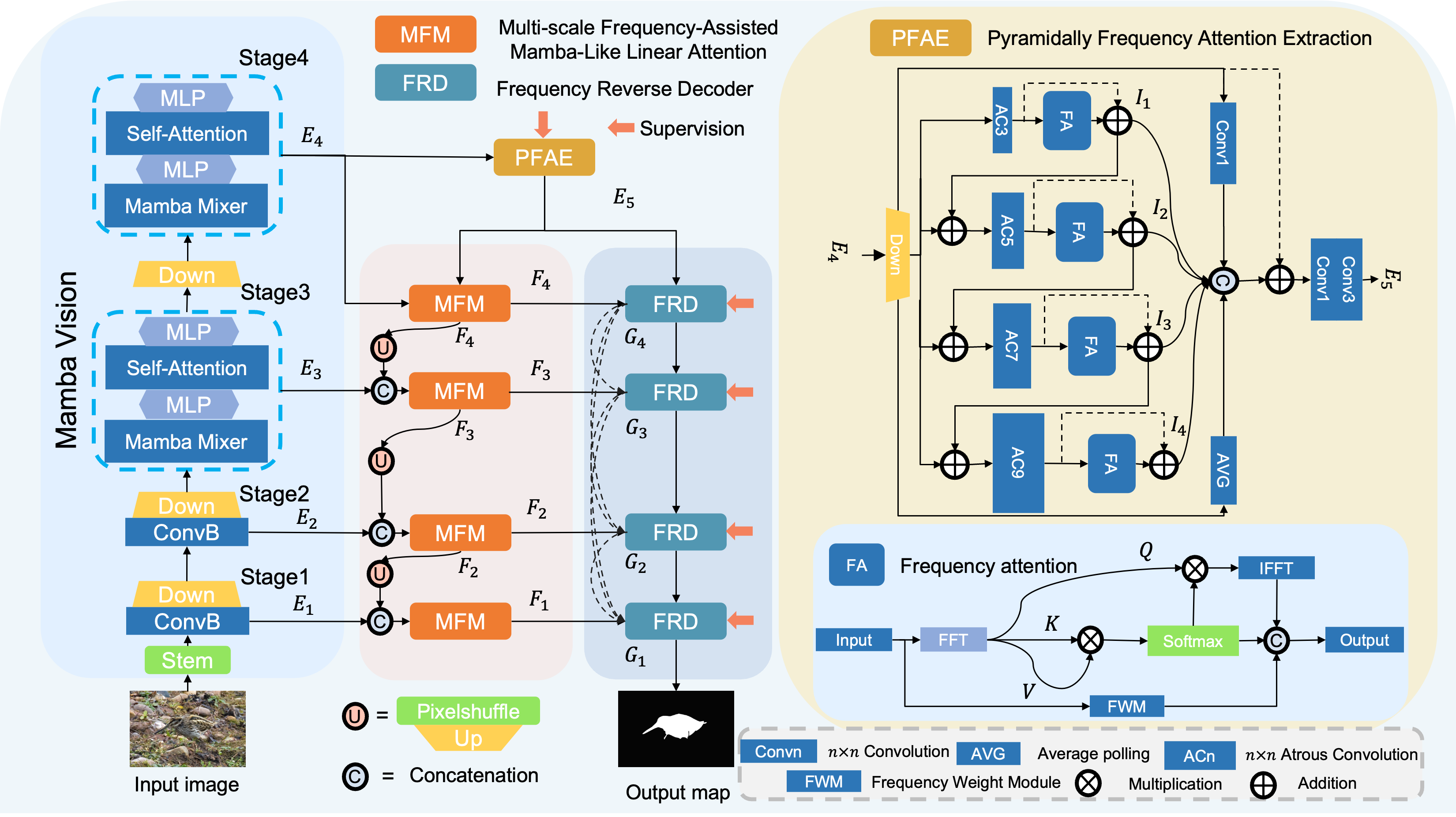}
    \caption{Overview of the proposed FMNet framework, proposed method adopts MambaVision as the encoder and introduces the following core modules: the Pyramidally Frequency Attention Extraction (PFAE) module based on frequency domain attention, the Multi-scale Frequency-Assisted Mamba-Like Linear Attention (MFM) module, and the Frequency Reverse Decoder (FRD).}
    \label{overview}
\end{figure*}

Transformer methods \cite{Vaswani2017AttentionIA}, with their ability to model long-range dependencies, have been widely applied in camouflaged object detection. However, the high computational cost and complex network structure of Transformers \cite{GLCONet}\cite{FSPNet} significantly limit their practicality.

Frequency-domain features \cite{FPNet}\cite{FSEL} have gained significant attention in camouflaged object detection due to their inherent global modeling capabilities, which effectively suppress background noise and improve the semantic clarity of camouflaged objects. This advantage is particularly evident in scenarios with blurred boundaries and occlusions, where camouflaged objects are difficult to distinguish. However, frequent transformations between frequency and spatial domains lead to increased computational complexity and parameter overhead.

In recent years, Mamba methods \cite{Gu2023MambaLS}\cite{Liu2024VMambaVS}, with their efficient attention mechanisms and lightweight design, have significantly reduced computational costs and shown great application potential. However, the potential of Mamba methods in camouflaged object detection remains underexplored.

Based on the above discussion, a novel method named Frequency-Assisted Mamba-Like Linear Attention Network (FMNet) is proposed. This method integrates feature information from both the frequency and spatial domains, introduces a multi-scale strategy to further extract global information, and leverages Mamba-Like Linear Attention (MLLA)\cite{mambalike} to optimize popular Transformer-based methods for COD. The comparison between our method and Transformer-based COD methods is illustrated in Fig. \ref{CT}. The main contributions in this paper can be summarized as follows.
% \begin{itemize}
%     \item 我们提出了区别于传统Transformer方法的FMNet，引入多尺度，频域辅助的方法提升伪装目标检测的性能
%     \item 我们设计了Multi-scale Frequency-Assisted Mamba-Like Linear Attention（MFM）模块，能够更全面的理解图像特征并减少了计算开销。
%     \item 为了增强图像的特征表示，我们引入了Pyramidally Frequency Attention Extraction（PAFE）模块和Frequency Reverse Decoder（FRD）模块
% \end{itemize} 
\begin{itemize}
\item FMNet is proposed, distinct from traditional Transformer methods, by incorporating multi-scale and frequency-assisted strategies to significantly enhance the performance of camouflaged object detection.
\item The Multi-scale Frequency-Assisted Mamba-Like Linear Attention (MFM) module is designed, which synergizes frequency-domain and spatial features, providing a more comprehensive understanding of image characteristics while effectively reducing computational complexity.
\item The Pyramidal Frequency Attention Extraction (PFAE) module and the Frequency Reverse Decoder (FRD) module are innovatively developed to enhance the representation of frequency-domain features and integrate multi-level information, further improving detection performance.
\end{itemize}

% \begin{figure*}[htbp]
% \centerline{\includegraphics{overview.png}}
% \caption{Example of a figure caption.}
% \label{fig}
% \end{figure*}

\section{Methodology}
%伪装目标在颜色、纹理和形状上与背景高度相似，其大小和形状在不同场景中可能存在显著变化，同时，其关键特征（如边缘和纹理）在实际环境中往往因遮挡或模糊而部分丢失。这些挑战使得仅依赖局部特征的传统检测方法难以在这些条件下准确提取目标信息。为了解决这些问题，通过频率域捕捉的全局特征与空间域提取的局部特征相互补充，从而增强整体检测能力。此外，所提出的网络架构改进了类似 Mamba 的线性注意力机制（Mamba-Like Linear Attention，MLLA），能够在多个尺度上高效提取局部和全局特征，同时有效降低复杂网络设计所带来的计算复杂性。
% Camouflaged objects exhibit a high degree of similarity to their background in terms of color, texture, and shape. Their size and shape can vary significantly across different scenarios, while their key features (e.g., edges and textures) are often partially lost in real-world environments due to occlusion or blurring. These challenges make it difficult for traditional detection methods relying solely on local features to accurately extract object information under such conditions. To address these issues, a complementary approach that combines global features captured in the frequency domain with local features extracted in the spatial domain is proposed, thereby enhancing overall detection capability. Moreover, the proposed network architecture improves the Mamba-Like Linear Attention (MLLA) mechanism, enabling efficient multi-scale extraction of both local and global features while significantly reducing the computational complexity associated with designing complex networks.

\subsection{Overview}

%  输入图像 \( I_c \in \mathbb{R}^{H \times W \times 3} \)，我们采用融合 Transformer 和 Mamba 架构的混合骨干网络，以高效提取初始特征\( \{E} = \{\{E}_i\}_{i=1}^4 \)，其分辨率为 \( \frac{W}{2^{i+1}} \times \frac{H}{2^{i+1}} \)的，随后，利用受 PAFE启发设计的PFFE提取多尺度的融合特征\( \{E}_5 \)，为了高效率的对跨长距离关系进行建模，引入FWMB，输出特征 \( \{F} = \{\{F}_i\}_{i=1}^4 \),最后经过FRD聚集多层次特征生成 \( \{F} = \{\{F}_i\}_{i=1}^4 \)

The proposed FMNet is shown in Fig. \ref{overview}. Given an input image $I_c \in \mathbb{R}^{H \times W \times 3}$, a hybrid backbone integrating Transformer and Mamba\cite{hatamizadeh2024mambavision} is adopted to efficiently extract the initial features $E_i$, where the resolution of each feature map is progressively reduced to $\frac{1}{2^{i+1}}$ of its original size. Inspired by previous works \cite{FSEL}\cite{PAFE}, a PFAE is employed to extract multi-scale fused features $E_5$. To efficiently model long-range dependencies across the feature domain, the MFM is used to refine the global context and produce optimized features $F_i$. Finally, FRD aggregates these multi-level features to produce the final feature maps $G_i$.

\begin{figure*}[!htbp]
    \centering
    \includegraphics[width=0.9\textwidth]{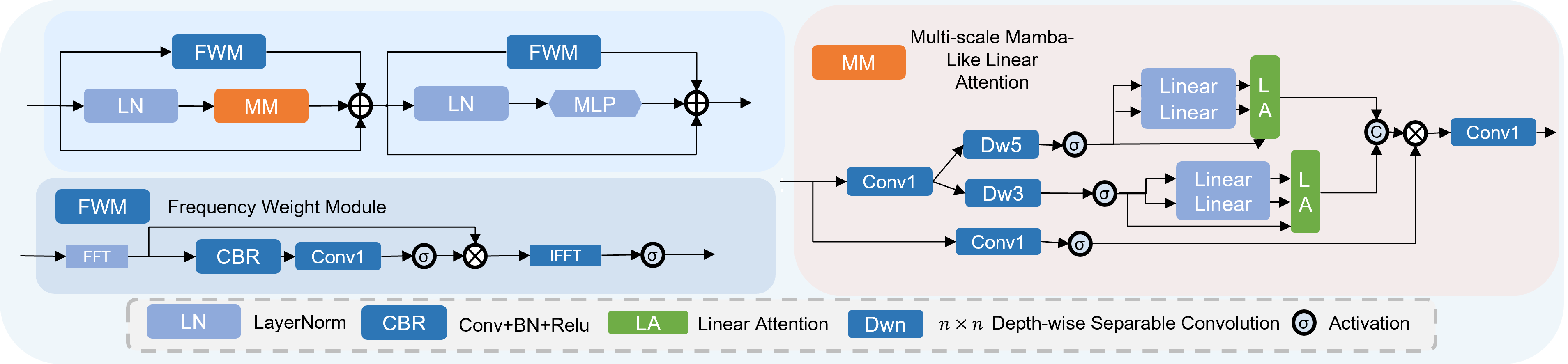}
    \caption{Details of the Multi-scale Frequency-Assisted Mamba-Like Linear Attention(MFM).}
    \label{MFM}
\end{figure*}   

\begin{figure}[htbp]
    \centering
    \includegraphics[width=0.45\textwidth]{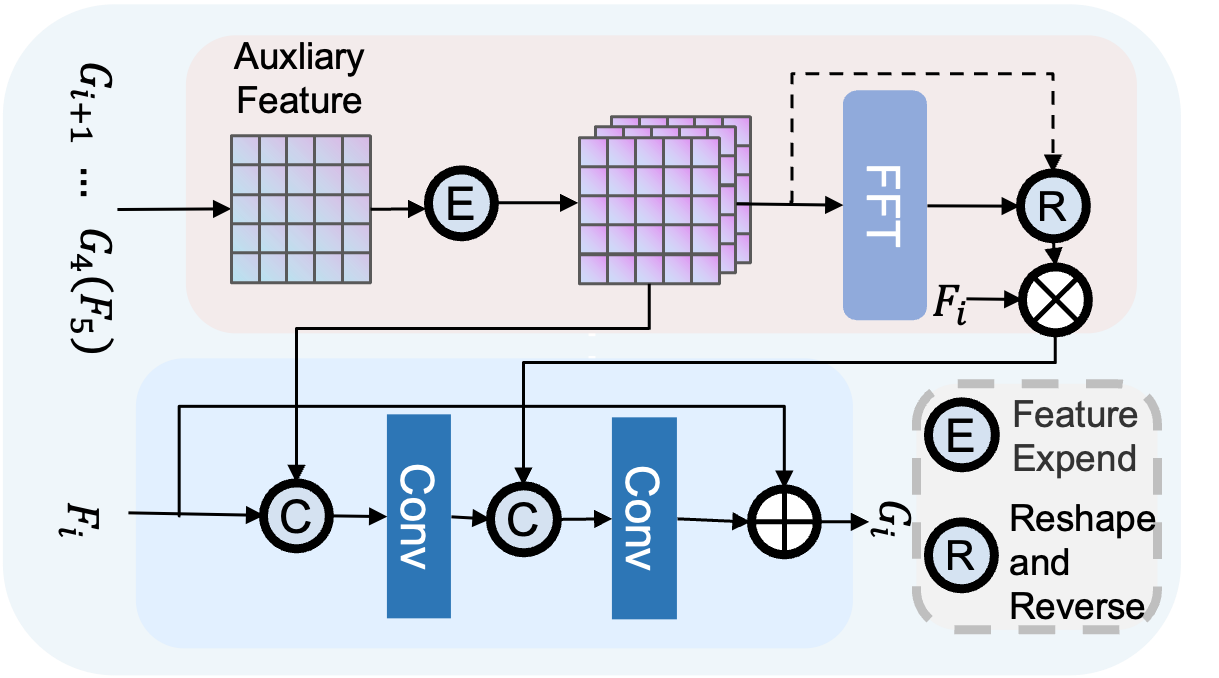}
    \caption{Details of Frequency Reverse Decoder.}
    \label{FRD}
\end{figure}

\subsection{Pyramidally Frequency Attention Extraction}\label{PFAE}
%为了解决单一感受野在捕获全局上下文信息和细粒度细节方面的局限性，许多受 Atrous Spatial Pyramid Pooling（ASPP）启发的模块已被引入到伪装目标检测（COD）网络中。然而，由于伪装目标在纹理、形状等方面与背景高度相似，仅依靠空间域卷积提取特征存在局限性。因此，与在空间域中结合注意力机制的 PAFE 模块不同，我们提出了一种频域注意力融合模块（FAFE），通过在频域中引入注意力机制，重点关注高频区域特征，从而更高效地提取诸如纹理和边缘等高频特征，弥补空间域在伪装目标检测中的不足。
The proposed PFAE module, as shown in Fig. \ref{overview}, integrates attention mechanisms in the frequency domain, allowing for more effective extraction of high-frequency features.
%具体来说，输入特征F_4 通过大小为1的卷积核减少通道数得到\hat{F_4}，接着加载四个膨胀率为z的空洞卷积层作为4个分支，每个分支包含一个频域注意力模块A_f，A_f

Specifically, the input feature $E_4$ is first passed through a $1 \times 1$ convolution to reduce the number of channels, resulting in $\hat{E_4}$. Subsequently, four dilated convolution layers with dilation rates $z$ are loaded as four branches:
\begin{equation}
    \tilde{I_n} = C_1AC_z(\hat{E_4}+I_{n-1}), \quad z = 2n-1, \quad n \geq 2
\end{equation}
each containing a frequency-domain attention module, this module uses the Fast Fourier Transform (FFT) to generate queries $Q$, keys $K$, and values $V$. After reshaping, the queries and keys are dot-multiplied to generate the transpose attention map $A'_f$:
\begin{equation}
Q, K, V = fft(\hat{E_4}),\quad A'_f = \tilde{Q} \odot \tilde{K}
\end{equation}
where $\hat{Q}$ and $\hat{K}$ are the results of applying the reshape operation to the query $Q$ and key $K$, since $A_f$ is of complex type, activate its real and imaginary parts separately, and then merge them:
\begin{equation}
A_f^{re} =\frac{A'_f + \text{coj}(A'_f)}{2}, \quad A_f^{im} = \frac{A'_f - \text{coj}(A'_f)}{2i}
\end{equation}
\begin{equation}
A_f = \Theta(\text{Sof}(A_f^{re}), \text{Sof}(A_f^{im}))
\end{equation}
where $\Theta(\cdot, \cdot)$ represents a function that combines the real and imaginary parts into a complex number, and $\text{Sof}(\cdot)$ represents the Softmax function. Subsequently, the attention map $A_f$ performs a dot product with $V$ to achieve weighted optimization, and the result is converted back to the original domain via the Inverse Fast Fourier Transform (IFFT). During this process, a Frequency Weight Module(FWM) is introduced, which will be discussed in Section \ref{D}, to perform residual connections. Finally, a $1 \times 1$ convolution is applied, and a residual connection with the features before the FFT transformation is added to generate the hybrid feature $I_n$:
\begin{equation}
I_n = C_1 \Phi( \| \text{ifft}(A_f \odot V) \|, FWM) + \tilde{I_n}
\end{equation}
where $\Phi$ represents the concatenation operation. The output $E_5$ is generated through convolution operations:
\begin{equation}
E_5 = C_3C_1(\Phi(J_1, J_2, J_3, J_4) + P_4^{128})
\end{equation}
where $C_k$ denotes a convolution operation with a kernel size of $k \times k$.

\subsection{Multi-scale Frequency-Assisted Mamba-Like Linear Attention}\label{D}
%多头自注意力模块（MultiHead Attention 模块）沿通道维度将每个查询、键和值分解为 N_h 个部分，生成查询 Q_i \in \mathbb{R}^{n \times D_h}、键 K_i \in \mathbb{R}^{n \times D_h} 和值 V_i \in \mathbb{R}^{n \times D_h},其中i表示第i个头，D_h是头的维度，每个头进行基于查询，键和值的自注意力学习，并连接所有输出，主要公式为：
The multi-head attention module\cite{Vaswani2017AttentionIA} decomposes input queries, keys, and values into \(N_h\) parts along the channel dimension, 
% where each part corresponds to a head. Specifically, the queries \(Q\), keys \(K\), and values \(V\) are split as follows:
% \begin{equation}
% Q_i \in \mathbb{R}^{n \times D_h}, \quad K_i \in \mathbb{R}^{n \times D_h}, \quad V_i \in \mathbb{R}^{n \times D_h}
% \end{equation}
% where \(i\) indicates the \(i\)-th head, \(D_h = \frac{D}{N_h}\) is the dimension of each head, and \(D\) is the total dimension of the channel. For each head, 
 self-attention learning is performed based on the queries, keys, and values using the formula:
\begin{equation}
A = \sigma\left(\frac{Q_i K_i^T}{\sqrt{D_h}}\right)V_i\label{shat}
\end{equation}
where $\sigma$ denotes the Sigmoid function, it can be seen that for an input feature map \(X \in \mathbb{R}^{H \times W \times D}\), the computational complexity is \(\mathcal{O}(H^2W^2D)\). Evidently, this complexity increases rapidly as the image resolution increases. As a result, many camouflaged object detection networks that adopt similar modules inevitably face high computational costs\cite{FSPNet}\cite{PFNet}. Inspired by methods aiming to address this issue\cite{linearattention}\cite{Gu2023MambaLS}, a module named Multi-scale Frequency-Assisted Mamba-Like Linear Attention (MFM) is proposed.

% \textbf{Wavelt down-sampled}. For an input feature map \(X \in \mathbb{R}^{H \times W \times D}\), we first apply a linear transformation using the embedding matrix \(W_d \in \mathbb{R}^{D \times \frac{D}{4}}\), reducing the channel dimension. This results in a transformed feature map \(\tilde{X} = X W_d\). Next, we perform a Discrete Wavelet Transform (DWT) on \(\tilde{X}\), decomposing it into four wavelet subbands (using Haar wavelets):

% \begin{equation}
% X_{LL}, X_{LH}, X_{HL}, X_{HH} \in \mathbb{R}^{\frac{H}{2} \times \frac{W}{2} \times \frac{D}{4}}
% \end{equation}
% Here, \(X_{LL}\) represents the low-frequency subband, capturing the basic structure of an object at a coarse-grained level, \(X_{LH}\), \(X_{HL}\), and \(X_{HH}\) represent the horizontal, vertical, and diagonal high-frequency subbands, preserving the texture details of the object.
% Each wavelet subband can be considered a down-sampled version of \(\tilde{X}\), covering all details of the feature map without loss of information.

% Next, the four wavelet subbands are concatenated along the channel dimension to \(\tilde{X} \in \mathbb{R}^{\frac{H}{2} \times \frac{W}{2} \times D}\). Then a convolution \(3 \times 3\) is applied to \(\tilde{X}\) to further capture local contextual information, resulting in the context-aware down-sampled feature map \(X^c\).

\textbf{Linearized attention}. Linear attention\cite{linearattention} replaces the non-linear Softmax function with linear normalization and introduces an additional kernel function $\phi$ into the query $q_j$ and key $k_k$. The output  $y_j$  can be implemented as an autoregressive model, which is expressed as follows:  
% \begin{equation}
%   y_j = \frac{Q_j \left(\sum_{k=1}^N K_k^\top V_k \right)}{Q_j \left(\sum_{k=1}^N K_k^\top \right)}\label{linearatt}  
% \end{equation}
% where $j$ and $k$ are the indices of the vectors. Equation \eqref{linearatt} can be implemented as an autoregressive model, which is expressed as follows:  
% \begin{equation}
% y_j = \frac{Q_j S_j}{Q_j Z_j}, \quad
% S_j = \sum_{k=1}^j K_k^\top V_k, \quad
% Z_j = \sum_{k=1}^j K_k^\top
% \end{equation} 
\begin{equation}
S_j = S_{j-1} + K_j^\top V_j, \quad
Z_j = Z_{j-1} + K_j^\top, \quad
y_j = \frac{Q_j S_j}{Q_j Z_j}\label{linearover}
\end{equation}

\begin{table*}[!htbp]
\centering
\caption{Comparison with State-of-the-Art Methods on three COD datasets. The symbols “↑”/“↓” indicate that higher/lower values are better. The best results are highlighted in \textcolor{red}{red}, and the second-best results are highlighted in \textcolor{blue}{blue}.}
\label{comparison}
\resizebox{\textwidth}{!}{
\begin{tabular}{l|c|cccc|cccc|cccc}
\toprule[1.5pt]
\multirow{2}{*}{\textbf{Model}} & \multirow{2}{*}{\textbf{Pub./Year}} & \multicolumn{4}{c|}{\textbf{CAMO-test}} & \multicolumn{4}{c|}{\textbf{COD10K-test}} & \multicolumn{4}{c}{\textbf{NC4K}} \\ 
               &                    & \cellcolor[HTML]{D3D3D3}$S_m \uparrow$ & \cellcolor[HTML]{D3D3D3}$F_\varphi \uparrow$ & \cellcolor[HTML]{D3D3D3}$E_m \uparrow$ & \cellcolor[HTML]{D3D3D3}$M \downarrow$ & \cellcolor[HTML]{D3D3D3}$S_m \uparrow$ & \cellcolor[HTML]{D3D3D3}$F_\varphi \uparrow$ & \cellcolor[HTML]{D3D3D3}$E_m \uparrow$ & \cellcolor[HTML]{D3D3D3}$M \downarrow$ & \cellcolor[HTML]{D3D3D3}$S_m \uparrow$ & \cellcolor[HTML]{D3D3D3}$F_\varphi \uparrow$ & \cellcolor[HTML]{D3D3D3}$E_m \uparrow$ & \cellcolor[HTML]{D3D3D3}$M \downarrow$ \\ 
\midrule[1.5pt]
JSOCOD\cite{JSOCOD}     & \textit{CVPR\textsubscript{21}} & 0.800 & 0.779 & 0.872 & 0.023 & 0.807 & 0.705 &0.882 & 0.035 & 0.841 & 0.803 & 0.966 & 0.047  \\ 
UGTR\cite{UGTR}     & \textit{ICCV\textsubscript{21}} & 0.785 & 0.738 & 0.823 & 0.086 & 0.818 & 0.712 & 0.853 & 0.035 & 0.839 & 0.787 & 0.874 & 0.052  \\ 
ZoomNet\cite{ZoomNet}    & \textit{CVPR\textsubscript{22}}  & 0.820 & 0.794 & 0.877 & 0.066 & 0.838 & 0.766 & 0.888 & 0.029 & 0.853 & 0.818 & 0.896 & 0.048 \\ 
SINet-V2\cite{COD10KV3}   & \textit{TPAMI\textsubscript{22}} & 0.820 & 0.782 & 0.882 & 0.070 & 0.815 & 0.718 & 0.887 & 0.037 & 0.847 & 0.805 & 0.903 & 0.048 \\ 
% FEDER\cite{FEDER}     & \textit{CVPR\textsubscript{23}} & 0.807 & 0.785 & 0.873 & 0.069 & 0.823 & 0.740 & 0.900 & 0.032 & 0.846 & 0.817 & 0.905 & 0.045 \\
FRINet\cite{FRINet}     & \textit{ACMMM\textsubscript{23}} & 0.865 & 0.848 & 0.924 & 0.046 & 0.864 & \textcolor{red}{0.810} & \textcolor{blue}{0.930} & 0.023 & 0.889 & \textcolor{blue}{0.866} & 0.937 & \textcolor{blue}{0.030} \\ 
% FSNet\cite{FSNet}       & \textit{TIP\textsubscript{23}}   & 0.880 & 0.861 & 0.933 & 0.041 & 0.870 & \textcolor{red}{0.810} & 0.927 & 0.023 & 0.891 & \textcolor{blue}{0.866} & 0.940 & 0.031 \\ 
FSPNet\cite{FSPNet}      & \textit{CVPR\textsubscript{23}}  & 0.857 & 0.830 & 0.899 & 0.050 & 0.851 & 0.769 & 0.895 & 0.026 & 0.879 & 0.843 & 0.915 & 0.035 \\ 

VSCode\cite{Luo2023VSCodeGV} & \textit{CVPR}\textsubscript{24} & 0.873 & 0.844  & 0.925 & 0.046 & \textcolor{blue}{0.869} & \textcolor{blue}{0.806} & 0.931 & \textcolor{blue}{0.023} & \textcolor{red}{0.891} & 0.863 & 0.935 & 0.032 \\ 
GLCONet\cite{GLCONet} & \textit{TNNLS\textsubscript{24}} & \textcolor{blue}{0.880} & \textcolor{red}{0.864} & \textcolor{blue}{0.940} & \textcolor{red}{0.038} & 0.860 & 0.784 & 0.929 & \textcolor{blue}{0.023} & 0.886 & 0.858 & \textcolor{blue}{0.942} & \textcolor{blue}{0.030} \\ 
% FSEL\cite{FSEL} & \textit{ECCV}\textsubscript{24} & \textcolor{blue}{0.885} & \textcolor{red}{0.864}  & \textcolor{blue}{0.942} & 0.040 & \textcolor{blue}{0.873} & 0.796 & 0.928 & \textcolor{blue}{0.021} & \textcolor{red}{0.892} & 0.864 & \textcolor{blue}{0.941} & \textcolor{blue}{0.030} \\ 
Ours & \textit{-} & \textcolor{red}{0.890} & \textcolor{blue}{0.863} & \textcolor{red}{0.946} & \textcolor{blue}{0.039} & \textcolor{red}{0.895} & 0.798 & \textcolor{red}{0.940} & \textcolor{red}{0.020} & \textcolor{blue}{0.890} & \textcolor{red}{0.871} & \textcolor{red}{0.944} & \textcolor{red}{0.029} \\ 
\bottomrule[1.5pt]
\end{tabular}
}
\end{table*}

\begin{table*}[!htbp]
\centering
\caption{Ablation analysis of the proposed method, with the best results highlighted in \textcolor{red}{red}.}
\label{ablation}
\begin{tabular}{l|cccc|c|c|cccc|cccc}
\toprule[1.5pt]
 \multirow{2}{*}{\textbf{No.}} & \multirow{2}{*}{\textbf{Baseline}} & \multirow{2}{*}{\textbf{PFAE}} & \multirow{2}{*}{\textbf{MFM}} & \multirow{2}{*}{\textbf{FRD}}  & \multirow{2}{*}{\textbf{Params (M)}} & \multirow{2}{*}{\textbf{FLOPs (G)}} & \multicolumn{4}{c|}{\textbf{COD10K-Test}} & \multicolumn{4}{c}{\textbf{NC4K}} \\ & & & &  &
               &                    &  \cellcolor[HTML]{D3D3D3}$S_m \uparrow$ & \cellcolor[HTML]{D3D3D3}$F_\varphi \uparrow$ & \cellcolor[HTML]{D3D3D3}$E_m \uparrow$ & \cellcolor[HTML]{D3D3D3}$M \downarrow$    &  \cellcolor[HTML]{D3D3D3}$S_m \uparrow$ & \cellcolor[HTML]{D3D3D3}$F_\varphi \uparrow$ & \cellcolor[HTML]{D3D3D3}$E_m \uparrow$ & \cellcolor[HTML]{D3D3D3}$M \downarrow$        \\  
\midrule[1.5pt]
(a) & \checkmark &  &  &                        & 50.14 & 29.11 & 0.823 & 0.702 & 0.894 & 0.035 & 0.855 & 0.801 & 0.913 & 0.044  \\ 
(b) & \checkmark & \checkmark &  &              & 50.97 & 30.37 & 0.831 & 0.713 & 0.892 & 0.033 & 0.860 & 0.821 & 0.914 & 0.043  \\ 
(c) & \checkmark &  & \checkmark &              & 62.10 & 41.45 & 0.881 & 0.756 & 0.922 & 0.025 & 0.887 & 0.846 & 0.930 & 0.034\\ 
(d) & \checkmark &  &  & \checkmark             & 53.01 & 31.76 & 0.865 & 0.742 & 0.907 & 0.030 & 0.875 & 0.842 & 0.919 & 0.038  \\ 
(e) & \checkmark & \checkmark & \checkmark &    & 62.93 & 42.75 & 0.883 & 0.779 & 0.919 & 0.024 & 0.885 & 0.852 & 0.929 & 0.032  \\ 
(f) & \checkmark & \checkmark &  & \checkmark   & 56.93 & 33.05 & 0.873 & 0.755 & 0.913 & 0.028 & 0.877 & 0.833 & 0.923 & 0.036  \\ 
(g) & \checkmark &  & \checkmark & \checkmark   & 68.06 & 44.13 & 0.893 & 0.787 & 0.936 & 0.022 & 0.883 & 0.865 & 0.942 & 0.030  \\ 
(h) & \checkmark & \checkmark & \checkmark & \checkmark & 68.89 & 45.39 & \textcolor{red}{0.895} & \textcolor{red}{0.798} & \textcolor{red}{0.940} & \textcolor{red}{0.020} & \textcolor{red}{0.890} & \textcolor{red}{0.871} & \textcolor{red}{0.944} & \textcolor{red}{0.029} \\ 
\bottomrule[1.5pt]
\end{tabular}
\end{table*}

\textbf{Selective state space model}. The classical state space model maps the input $ x(t) \in \mathbb{R}$  to the output $ y(t) \in \mathbb{R} $ through the hidden state $ h(t) \in \mathbb{R}^{d \times 1} $. Mamba\cite{Gu2023MambaLS} introduces the selective state space model, which discretizes the parameters using zero-order hold (ZOH). MLLA\cite{mambalike} rewrites Mamba into the following equivalent form:
% :
% \begin{equation}
%     \overline{A}_j = \exp(\Delta_j A), \quad \overline{B}_j = \Delta_j B_j
% \end{equation}
% \begin{equation}
% h_j = \overline{A}_j h_{j-1} + \overline{B}_j x_j, \quad y_j = C_j h_j + Dx_j\label{mamba}
% \end{equation}  
% where $x_j$ is the input, and the parameters $B_j, C_j, \Delta_j$ are functions of $x_j$. Through (1) $\overline{A}_i h_{i-1} = \tilde{A}i \odot h_{i-1}$, $\widetilde{A}_j = \text{diag}(\overline{A}_j) \in \mathbb{R}^{d \times 1}$, 
% (2) converting $B_j = \Delta_j B_j$ and $\Delta_j \in \mathbb{R}$ to $\overline{B}_j x_j = \Delta_j B_j x_j = B_j (\Delta_j x_j) = B_j (\Delta_j \odot x_j)$, and (3) $Dx_j = D \odot x_j$, equation \eqref{mamba} can be rewritten as:
\begin{equation}
h_j = \widetilde{A}_j \odot h_{j-1} + B_j (\Delta_j \odot x_j), \quad y_j = C_j h_j + D \odot x_j\label{mambaover}
\end{equation}
where $\odot$ denotes the element-wise multiplication, $B_j, C_j, \Delta_j$ are derived from the input, $x_j, \Delta_j \in \mathbb{R}^{1 \times C}$, and $y_j \in \mathbb{R}^{1 \times C}$.

It can be observed that \eqref{linearover} and \eqref{mambaover} have a close relationship, specifically: $h_j \sim S_j \in \mathbb{R}^{d \times C}, B_j \sim K_j^\top \in \mathbb{R}^{d \times 1}, x_j \sim V_j \in \mathbb{R}^{1 \times d},  C_j \sim Q_j \in \mathbb{R}^{1 \times d}$, $\tilde{A}_i$ plays the role of a forget gate in the Selective State Space Model (SSM). The MLLA module inherits the advantages of Mamba, adopting a structure similar to Mamba and introducing the forget gate mechanism.
At the same time, to better adapt to vision tasks, MLLA replaces the traditional forget gate with positional encodings LePE\cite{lepe}, RoPE\cite{Su2021RoFormerET} and CPE\cite{cpe}:
\begin{equation}
LePE(x) = x + DWConv(x) W_L  
\end{equation}
\begin{equation}
RoPE(x_m, \theta_i) = x_m \cdot (\cos(m\theta_i) + \sin(m\theta_i)) 
\end{equation}
\begin{equation}
L_a = Att(RoPE(Q), RoPE(K), V + LePE(V)) 
\end{equation}
where $W_L$ is a learnable weight matrix, $DWConv$ denotes a depth-wise convolution, $x_m$ is the $m$-th dimension of the input, and $\theta_i$ is the angle related to the position, the complexity of the Linear Attention module is $O(NCd) = O(HWDD_h)$.

\textbf{Frequency-assisted Multi-scale MLLA}. %MLLA的优化让我们设计类似\cite{multiscaletransfomer}的多尺度结构的复杂度能够接受，输入特征$E_i$首先经过层归一化得到张量\tilde{E_i},沿通道纬度平分后经过$1 \times 1$卷积后跟着$n \times n$的深度卷积生成不同尺度的张量，经过Linear Attention层后拼接在一起，整体公式可以表示为：
The proposed MFM module is shown in Fig. \ref{MFM}, where the figure omits reshaping operations for clarity. The optimization of MLLA allows us to design a multi-scale structure similar to \cite{mutiscaletransfomer} with acceptable complexity. The input feature $E_i$ is first processed through CPE and layer normalization to obtain the tensor $\tilde{E_i}$. Then, it is split along the channel dimension and processed through $1 \times 1$ convolutions, followed by $n \times n$ depth-wise convolutions to generate tensors of different scales. These tensors are concatenated together after passing through the Linear Attention layer. The overall process is represented by the following formula:
\begin{equation}
    A_i^n = L_a(R(\sigma(D_nC_1(\tilde{E_i})))
\end{equation}
\begin{equation}
    A_i = \Phi(A_i^3,A_i^5)\quad F_i^1 = L(A_i \odot R(\sigma(C_1(\tilde{E_i})))
\end{equation}
where $R$  represents the reshape operation,  $D_n$  denotes the depthwise convolution kernel,  $C_1$  refers to the convolution kernel with a size of 1, $\Phi$ represents the concatenation operation, and  $F_i^1$  is the feature map output from the first stage.

A Frequency Weight Module (FWM) to enhance the representation of frequency-domain information through frequency residual connections is also designed:
\begin{equation}
   FWM =  \| ifft(W(fft(X)) * fft(X)) \| 
\end{equation}
where $W(\cdot)$  represents a series of operations consisting of convolution, batch normalization, GELU, convolution, and sigmoid functions applied sequentially, after passing through the multi-scale MLLA module,  $F_i^2$  is obtained:
\begin{equation}
    F_i^2 = CPE(F_i^1 + FWM(E_i) + E_i)
\end{equation}
Finally, the overall output  $F_i$  of the MFM can be obtained using the following formula: 
\begin{equation}
    F_i = F_i^2 + LN(Mlp(F_i^2)) + FWM(F_i^2)
\end{equation}
where $LN$ refers to Layer Normalization.

\begin{figure*}[!htbp]
    \centering
    \includegraphics[width=\textwidth]{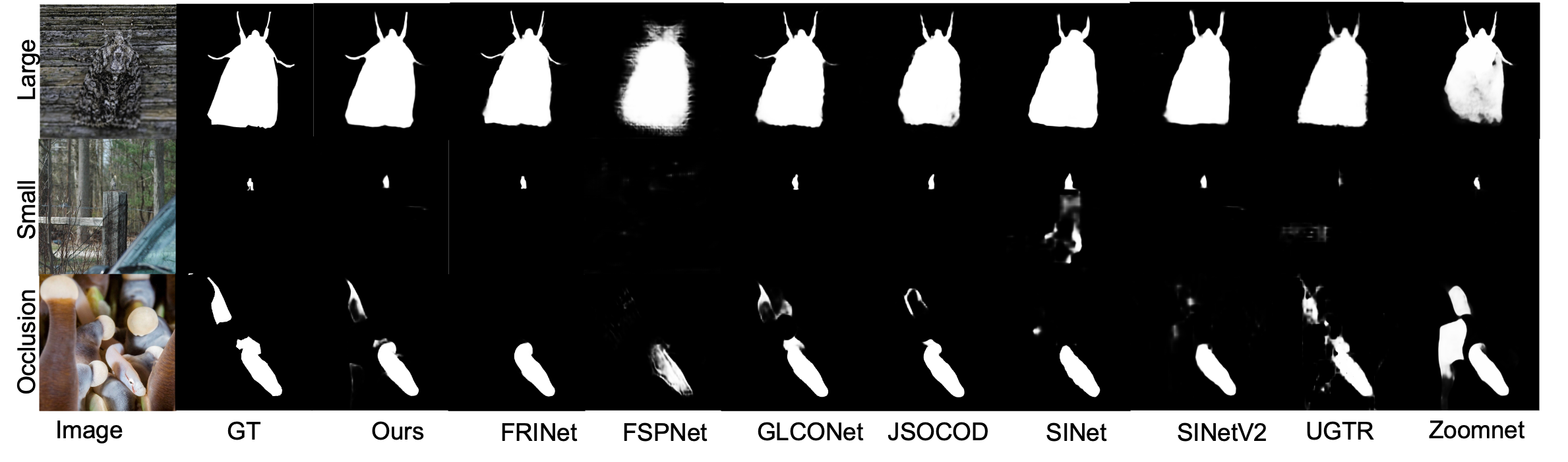}
    \caption{Visual comparison of the prediction maps with state-of-the-art methods, including large objects, small objects, and occluded objects.}
    \label{VIS}
\end{figure*}

% \begin{figure}[!htbp]
%     \centering
%     \includegraphics[width=0.45\textwidth]{VIS.png}
%     \caption{Visual comparison of the prediction maps with state-of-the-art methods, including large objects, small objects, and occluded objects.}
%     \label{VIS}
% \end{figure}

\subsection{Frequency Reverse Decoder}
The FRD module is shown in Fig. \ref{FRD} Unlike existing COD methods \cite{codassplike3} \cite{FSPNet}.
%FRD的输入为两个部分：辅助输入和主输入，FRD的输入为两个部分：辅助输入和主输入，将来自高层的特征图G_i(F_5)作为辅助输入，首先通过双线插值和纬度扩展确保辅助输入的尺寸和通道数

The input of the FRD consists of two parts: the auxiliary input and the main input $F_i$. The high-level feature map $G_{i+1}(F_5)$ is used as an auxiliary input. First, bilinear interpolation and channel dimension expansion are applied to ensure that the auxiliary input matches the size and number of channels of the main input. Subsequently, the auxiliary input is concatenated with the main input along the channel dimension to obtain the feature $G_i^1$:
\begin{equation}
    G_i^1 = \Phi (F_i,Ex(G_{i+1}),...,Ex(Z)), \quad Z = \{G_4,F_5\}
\end{equation}
the reverse attention is applied to the auxiliary features to generate a frequency-spatial hybrid reverse attention map $RA$, which is then used to produce the frequency-optimized reverse feature $G_i^2$:
\begin{equation}
RA =\sum{(1 - \sigma(G_{i+1})) + (1 - \sigma( \| fft(G_{i+1}) \|))}
\end{equation}
\begin{equation}
    G_i^2 = RA*F_i
\end{equation}
finally, $G_i^1$ undergoes a series of convolution operations and is then integrated with $G_i^2$ to generate the final feature $G_i$, $G_i$ will serve as the auxiliary input for the next layer’s FRD.
\begin{equation}
    G_i = C_3\Phi(G_i^1,Con(G_i^2))+F_i
\end{equation}

\subsection{Loss Function}
Similar to these methods\cite{codassplike3}\cite{FSPNet}, the weighted binary cross-entropy (BCE) and weighted intersection over union (IoU)\cite{IOU} are used as loss functions to supervise the multi-level features $G_i$. The loss function is defined as follows:
\begin{equation}
    \mathcal{L}_{\text{all}} = \sum_{i=1}^{5} 2^{1-i} \left( \mathcal{L}_{\text{bce}}^w(G_i, GT) + \mathcal{L}_{\text{iou}}^w(G_i, GT) \right)
\end{equation}
where $\mathcal{L}_{\text{bce}}^w$ and $\mathcal{L}_{\text{iou}}^w$ represent the weighted BCE and IoU loss functions, respectively.

% \begin{table*}[ht]
% \centering
% \caption{Efficiency analysis of our FMNet and existing COD methods.}
% \label{efficiency}
% \resizebox{\textwidth}{!}{
% \begin{tabular}{l|cccccccccc}
% \toprule[1.5pt]
% & FEDER\cite{FEDER} & ZoomNet\cite{ZoomNet} & PFNet\cite{PFNet}  & MGL\cite{MGL}  & UGTR\cite{UGTR}  & JSOCOD\cite{JSOCOD}   & FSEL\cite{FSEL} & FSPNet\cite{FSPNet}   & GLCONet \cite{GLCONet} & Ours \\ 
% \midrule[1.5pt]
% Parameters (M) & 37.37 & \textcolor{red}{32.38} & 46.50 & 63.60 & 48.87 & 217.98  &  67.13 & 273.79  & 91.35 & 68.89\\ 
% FLOPs (G)      & \textcolor{red}{23.98} & 203.50 & 53.22 & 553.94 & 1000.01 & 112.34  & 54.73 & 283.31  & 173.58 & 45.39\\ 
% \bottomrule[1.5pt]
% \end{tabular}
% }
% \end{table*}

\begin{table}[htbp]
\caption{Efficiency analysis of the proposed FMNet and existing COD methods, P represents the parameters and F represents the FLOPs. }
\label{efficiency}
\begin{center}
\begin{tabular}{l|ccccccccc}
\toprule[1.5pt]
    & JSOCOD\cite{JSOCOD}   & FSPNet\cite{FSPNet}    & GLCONet \cite{GLCONet} & Ours \\ 
\midrule[1.5pt]
P (M)  & 217.98  &  273.79   & 91.35 & \textcolor{red}{68.89}\\ 
F (G)      & 112.34  &  283.31  & 173.58 & \textcolor{red}{45.39}\\ 
\bottomrule[1.5pt]
\end{tabular}
\end{center}
\end{table}

\section{Experiments}
\subsection{Experiment Setting}
The FMNet method is evaluated on the CAMO\cite{CAMO}, COD10K\cite{COD10K}, and NC4K\cite{Nc4k} datasets, using 4,000 images from the CAMO\cite{CAMO} and COD10K\cite{COD10K} dataset for training. The FMNet method is trained on three NVIDIA GTX 4090 GPUs (24GB) and utilizes the pre-trained MambaVision as an encoder to extract initial features. Input images are resized to $416 \times 416$ and processed with data augmentation techniques such as random horizontal flipping and cropping. During training, the batch size is set to 30, over 100 epochs. The initial learning rate is set to $1e^{-4}$, and the Adam optimizer is used, with the learning rate reduced by a factor of 10 every 50 epochs. To evaluate method performance, the following metrics are adopted: S-measure ($S_m$), E-measure ($E_m$), Average F-measure ($F_\varphi$), and Mean Absolute Error ($\mathcal{M}$).
\subsection{Comparisons with the State-of-the-Art Methods}
The FMNet is compared with 8 state-of-the-art methods, including JSOCOD\cite{JSOCOD}, UGTR\cite{UGTR}, ZoomNet\cite{ZoomNet}, SINet-V2\cite{COD10KV3}, FRINet\cite{FRINet}, FSPNet\cite{FSPNet}, VSCode\cite{Luo2023VSCodeGV} and GLCONet\cite{GLCONet}. All prediction results in this paper were provided by the original authors or obtained from open-source codes.

The quantitative comparison between the FMNet method and other SOTA methods is summarized in Tab. \ref{comparison}. The results show that the method outperforms other methods in various evaluation metrics, demonstrating exceptional performance. Additionally, the parameters and FLOPs comparison between FMNet and other methods is provided in Tab. \ref{efficiency}. It can be seen that the proposed method demonstrates an efficiency advantage compared to complex networks.

Visual comparisons between FMNet and other methods are presented in Fig. \ref{VIS}, including segmentation results for large objects, small objects, and occluded objects. The results indicate that the proposed method demonstrates robust performance in all scenarios.

\subsection{Ablation Study}
The quantitative results of the components are presented in the FMNet method in Table \ref{ablation}. Specifically, the effectiveness of PFAE, MFM, and FRD is validated, as well as their impact on parameters and FLOPs. The experimental results show that adding the proposed modules to the baseline leads to a significant improvement in prediction performance. From Tab. \ref{ablation}(e), Tab. \ref{ablation}(f), and Tab. \ref{ablation}(g), it can be observed that these three components exhibit good compatibility with each other.

Additionally, the efficiency ablation study reveals that the MLLA module consumes the most parameters and FLOPs while contributing the most to performance improvement, further proving that it is the most critical part for optimization.

% \begin{figure}[htbp]
% \centerline{\includegraphics{fig1.png}}
% \caption{Example of a figure caption.}
% \label{fig}
% \end{figure}

\section{Conclusion}
This paper proposes a novel FMNet method for camouflaged object detection, with its core centered on a high-performance multi-scale frequency-assisted module (MFM). FMNet integrates feature information from both the spatial and frequency domains, achieving more precise object segmentation. Furthermore, a Pyramidally Frequency Attention Extraction (PFAE) module is designed to extract multi-scale features, and a Frequency Reverse Decoder (FRD) to cross-layer aggregation and reverse optimization, further improving the model performance. Extensive comparative experiments demonstrate that FMNet outperforms existing SOTA methods on multiple benchmark datasets. In the future, we will improve FMNet’s adaptability to complex environments and optimize it for edge device deployment to enhance its practical applications.

\section*{Acknowledgments}
This work is supported by the National Natural Science Foundation of China (No. 62172267), the Project of Key Laboratory of Silicate Cultural Relics Conservation (Shanghai University), Ministry of Education (No. SCRC2023ZZ02ZD).

\vspace{12pt}

\end{document}